\documentclass{article}

\usepackage{color}
\usepackage[final]{nips_2016}

\usepackage[utf8]{inputenc} 
\usepackage[T1]{fontenc}    
\usepackage{hyperref}       
\usepackage{url}            
\usepackage{booktabs}       
\usepackage{amsfonts}       
\usepackage{amsmath}       
\usepackage{nicefrac}       
\usepackage{microtype}      
\usepackage{graphicx}
\usepackage{algorithm}
\usepackage{algorithmic}

\title{Spatial contrasting for deep unsupervised learning}

\author{
  Elad Hoffer\\
  Technion - Israel Institute of Technology\\
  Haifa, Israel\\
  \texttt{ehoffer@tx.technion.ac.il} \\
    \And
  Itay Hubara\\
  Technion - Israel Institute of Technology\\
  Haifa, Israel\\
  \texttt{itayh@tx.technion.ac.il} \\
      \And
  Nir Ailon\\
  Technion - Israel Institute of Technology\\
  Haifa, Israel\\
  \texttt{nailon@cs.technion.ac.il}\\
}

\begin{document}

\maketitle

\begin{abstract}
Convolutional networks have marked their place over the last few years as the best performing model for various visual tasks. 
They are, however, most suited for supervised learning from large amounts of labeled data. 
Previous attempts have been made to use unlabeled data to improve model performance by applying unsupervised techniques. These attempts require different architectures and training methods. In this work we present a novel approach for unsupervised training of
Convolutional networks that is based on contrasting between spatial regions within images. 
This criterion can be employed within conventional neural networks and trained using standard techniques such as  SGD and back-propagation, thus complementing supervised methods.
\end{abstract}

\section{Introduction}
For the past few years convolutional networks (ConvNets, CNNs)  \citet{lecun1998gradient}  have proven themselves as a successful model
for vision related tasks \citet{Krizhevsky2012} \citet{mnih2015human} \citet{pinheiro2015learning} \citet{razavian2014cnn}. 
A convolutional network is composed of multiple convolutional and pooling layers,
followed by a fully-connected affine transformations. As with other neural network models, each layer is typically followed by
a non-linearity transformation such as a rectified-linear unit (ReLU). \\
A convolutional layer is applied by cross correlating an image with a trainable weight filter. This stems from the
assumption of stationarity in natural images, which means that  features learned for one local region in an image can be shared for other regions and images. 

Deep learning models, including convolutional networks, are usually trained in a supervised manner, requiring large amounts
of labeled data (ranging between thousands to millions of examples per-class for classification tasks) in almost all modern applications. 
These models are optimized a variant of stochastic-gradient-descent (SGD) over batches of images sampled from the
whole training dataset and their ground truth-labels. Gradient estimation for each one of the optimized parameters is done by back propagating the
objective error from the final layer towards the input. This is commonly known as "backpropagation" \citet{rumelhart1988learning}.

One early well known usage of unsupervised training of deep architectures was as part of a pre-training procedure used for obtaining an effective initial state
of the model. The network was later fine-tuned in a supervised manner as displayed by \citet{hinton2007recognize}.  
Such unsupervised pre-training procedures were later abandoned, since they provided no apparent benefit over other initialization heuristics in 
more careful fully supervised training regimes. This led to the de-facto almost exclusive usage of neural networks in supervised environments.

In this work we will present a novel unsupervised learning criterion for convolutional network based on comparison of features extracted from regions within images.
Our experiments indicate that by using this criterion to pre-train networks we can improve their performance and achieve state-of-the-art results.

\section{Problems with Current Approaches}


The majority of unsupervised optimization criteria currently used are based on variations of reconstruction losses.  
One  limitation of this fact is that a pixel level reconstruction is non-compliant with the idea  of a discriminative objective, 
which is expected to be agnostic to low level information in the input.
In addition, it is evident that MSE is not best suited as a measurement to
compare images, for example, viewing the possibly large square-error between an image and a single pixel shifted copy of it. 
Another problem  with recent approaches such as \citet{rasmus2015semi,zeiler2010deconvolutional}  is their need to extensively modify the original convolutional network model.
This leads to a gap between unsupervised method and the state-of-the-art, supervised, models for classification - which can hurt future attempt to reconcile them in a unified framework, and also to efficiently leverage unlabeled data with otherwise supervised regimes.

\section{Learning by Comparisons}
The most common way to train NN is by defining a loss function between the target values and the network output. Learning by comparison approaches the supervised task from a different angle. The main idea is to use distance comparisons between samples to learn useful representations. For example, we consider relative and qualitative examples of the form “$X_1$ is closer to $X_2$ than $X_1$ is to $X_3$. Using a comparative measure with neural network to learn embedding space was introduced in the ``Siamese network'' framework by \citet{bromley1993signature} and later used in the works of \citet{chopra2005learning}.
One use for this methods is when the number of classes is  too large or expected to vary over time, as in the case
of face verification, where a face contained in an image has to compared against another image of a face. 

\section{Our Contribution: Spatial Contrasting}
One implicit assumption in convolutional networks, is that features 
are gradually learned hierarchically, each level in the hierarchy corresponding to a layer in the network.
Each spatial location within a layer corresponds to a
region in the original image.  It is empirically observed that deeper layers tend to contain more `abstract' information from the image. 
Intuitively,
features describing 
different regions within the same image are likely to be  semantically similar (e.g.  different parts of an animal), and
indeed the corresponding deep representations tend to be similar.
Conversely, regions from two probably unrelated images (say, two images chosen at
random) tend to be far from each other in the deep representation.
This logic is commonly used in modern deep networks such as \citet{inception} \citet{nin} \citet{res}, 
where a global average pooling is used to aggregate spatial features in the final layer used for classification. 

Our suggestion is that this property, often observed as a side effect of supervised applications, can be used as a desired objective when learning deep representations
 in an unsupervised task.  Later, the resulting representation can be used, as typically done, as a starting point
or a supervised learning task.
We call this idea which we formalize below \emph{Spatial contrasting}.  
The spatial contrasting criterion is similar to noise contrasting estimation \cite{nce1} \cite{nce2}, in trying to train a model by
maximizing the expected probability on desired inputs, while minimizing it on contrasting sampled measurements.

\subsection{Formulation}
We will concern ourselves with samples of images patches $\tilde{x}^{(m)}$ taken from an image $x$. Our convolutional network model, denoted by $F(x)$, extracts spatial features $f$ so that $f^{(m)}=F(\tilde{x}^{(m)})$ for an image patch $\tilde{x}^{(m)}$. We wish to optimize our
model such that for two features representing patches taken from the same image $\tilde{x}_i^{(1)},\tilde{x}_i^{(2)}\in x_i$
for which $f_i^{(1)}=F(\tilde{x}_i^{(1)})$ and $f_i^{(2)}=F(\tilde{x}_i^{(2)})$, 
the conditional probability $P(f_i^{(1)}|f_i^{(2)})$ will be maximized. \\
This means that features from a patch taken from a specific image  can  effectively predict,
under our model, features extracted from other patches in the same image.  Conversely, we want our model to minimize $P(f_i|f_j)$ for $i,j$ being two patches taken
from distinct images.
Following the logic presented before, we will need to sample
\emph{contrasting patch} $\tilde{x}^{(1)}_j$ from a different image $x_j$ such that $P(f_i^{(1)}|f_i^{(2)})>P(f_j^{(1)}|f_i^{(2)})$, where $f_j^{(1)}=F(\tilde{x}^{(1)}_j)$. 
In order to obtain contrasting samples,
we use regions from two random images in the training set.
We will use a distance ratio, described earlier \ref{ratio_distance} for the supervised case, to represent the probability two feature vectors were taken from the same image.
The resulting training loss for a pair of  images will be defined as

\begin{equation}
 L_{SC}(x_1,x_2) = -\log \frac{e^{-\|f_1^{(1)} - f_1^{(2)}\|_2}}{e^{-\|f_1^{(1)} - f_1^{(2)}\|_2} + e^{-\|f_1^{(1)} - f_2^{(1)}\|_2}}
\end{equation}

Effectively minimizing a log-probability under the SoftMax measure.


\subsection{Method}
Since training convolutional network is done in batches of images, we can use the multiple samples in each batch to train our model. Each image serves as a source for
both an anchor and positive patches, for which the corresponding features should be closer, and also a source for contrasting samples for all the other images in that batch.
For a batch of $N$ images, two samples from each image are taken, and $N^2$ 
different distance comparisons are made. The final loss is the average distance ratio
for images in the batch:
\begin{equation}
\bar{L_{SC}}(\{x\}_{i=1}^N) = \frac{1}{N}\sum_{i=1}^N L_{SC}(x_i,\{x\}_{j\ne i}) = -\frac{1}{N}\sum_{i=1}^N \log \frac{e^{-\|f_i^{(1)} - f_i^{(2)}\|_2}}{\sum_{j=1}^N e^{-\|f_i^{(1)} - f_j^{(2)}\|_2}}
\end{equation}
Since the criterion is differentiable with respect to its inputs, it is fully compliant with standard methods for training convolutional network and specifically using
backpropagation and gradient descent.
Furthermore, SC can be applied to any layer in the network hierarchy.
In fact, SC can be used at  multiple layers within 
the same convolutional network.

\section{Experiments}
In this section we report empirical results showing that using SC loss as an unsupervised pretraining procedure 
can improve state-of-the-art performance on subsequent classification. 
In each one of the experiments, we used the spatial contrasting criterion to train the network on the unlabeled images. 
We then used the trained model as an initialization for a supervised training on the complete labeled dataset.
\subsection{Results on STL10}
This dataset consists of $100,000 \ 96\times 96$ colored, unlabeled images, together with 
another set of $5,000$ labeled training images and $8,000$ test images .  The label space consists of 10 object classes. 

\begin{table}[t]
  \caption{State of the art results on STL-10 dataset}
  \label{stl10_table}
  \centering
  \begin{tabular}{lll}
    \toprule
    \bf Model     & \bf STL-10 test accuracy \\
    \midrule
 Zero-bias Convnets - \citet{paine2014analysis}                        &$70.2\%$\\
 Triplet network - \citet{hoffer2015deep} 	                       &$70.7 \%$ \\
 Exemplar Convnets - \citet{dosovitskiy2014discriminative}             &$72.8\%$ \\
 Target Coding - \citet{yang2015deep}				       &$73.15\%$ \\
 Stacked what-where AE - \citet{whatwhereae}			       &$74.33\%$ \\
   \midrule
Spatial contrasting initialization (this work)                           	    &\bf $81.34\% \pm 0.1$ \\          
The same model without initialization                        	 		   &\bf $72.6\% \pm 0.1$ \\     
    \bottomrule
  \end{tabular}
\end{table}

\subsection{Results on Cifar10}
The well known CIFAR-10 is an image classification benchmark dataset containing $50,000$ training images and $10,000$ test images.  The image sizes  $32 \times 32$ pixels, with color.  The classes are
airplanes, automobiles, birds, cats, deer, dogs, frogs, horses, ships and trucks
For Cifar10, we used a previously used setting  \citet{coates2012learning} \citet{hui2013direct}  \citet{dosovitskiy2014discriminative} to test a model's ability to learn from
unlabeled images. In this setting, only $4,000$ samples from the available $50,000$ are used with their label annotation, but the entire dataset is used for unsupervised learning.

      \begin{table}[t]
  \caption{State of the art results on Cifar10 dataset with only 4000 labeled samples}
  \label{cifar10_table}
  \centering
  \begin{tabular}{lll}
    \toprule
    \bf Model     & \bf Cifar10 (400 per class) test accuracy \\
    \midrule
 Convolutional K-means Network - \citet{coates2012learning}                  &$70.7\%$\\
 View-Invariant K-means - \citet{hui2013direct}                            &$72.6\%$\\
 DCGAN - \citet{radford2015unsupervised} 	                      		 &$73.8 \%$ \\
 Exemplar Convnets - \citet{dosovitskiy2014discriminative}             &$76.6\%$ \\
 Ladder networks - \citet{rasmus2015semi}				       &$79.6\%$ \\

   \midrule
Spatial contrasting initialization (this work)                         &\bf $79.2\% \pm 0.3$ \\          
The same model without initialization                        	       &\bf $72.4\% \pm 0.1$ \\     
    \bottomrule
  \end{tabular}
\end{table}

\section{Conclusions and future work}
In this work we presented spatial contrasting - a novel unsupervised criterion for training convolutional networks on unlabeled data. Its is based on comparison between spatial features
sampled from a number of images. We've shown empirically that using spatial contrasting as a pretraining technique to initialize a ConvNet, 
can improve its performance on a subsequent supervised training. 
In cases where a lot of unlabeled data is available, such as the STL10 dataset, this translates to state-of-the-art classification accuracy in the final model.

Since the spatial contrasting loss is a differentiable estimation that can be computed within a network in parallel to supervised losses,
future work will attempt to embed it as a semi-supervised model. This usage will allow to create models that can leverage
both labeled an unlabeled data, and can be compared to similar semi-supervised models such as the ladder network \citet{rasmus2015semi}.
It is is also apparent that contrasting can occur in dimensions other than the spatial, the most straightforward is the temporal one. This suggests that similar
training procedure can be applied on segments of sequences to learn useful representation without explicit supervision.

\small
\bibliographystyle{plainnat}
\bibliography{spatialcontrasting}

\end{document}